\documentclass[11pt]{article}

\usepackage[final]{acl}

\usepackage{times}
\usepackage{latexsym}

\usepackage[T1]{fontenc}

\usepackage[utf8]{inputenc}

\usepackage{microtype}

\usepackage{inconsolata}

\usepackage{graphicx}

\usepackage{xspace}
\usepackage{CJKutf8}
\usepackage{amssymb}
\usepackage{soul}
\usepackage{multirow}
\usepackage{float}
\usepackage{graphicx}
\usepackage{longtable}
\usepackage{makecell}
\usepackage{booktabs}
\usepackage{arydshln}
\usepackage{tabularx}
\usepackage{stfloats}
\usepackage{enumitem}
\usepackage{cellspace}
\usepackage{amsmath}
\usepackage{threeparttable}
\usepackage{subfigure}
\usepackage[tone]{tipa}
\usepackage{CJKutf8}

\author{Ruyuan Wan\textsuperscript{1}~~~~Changye Li\textsuperscript{2}~~~~Ting-Hao `Kenneth' Huang\textsuperscript{1}\\
\textsuperscript{1}The Pennsylvania State University \quad     \textsuperscript{2}University of Washington\\
\textsuperscript{1}\texttt{\{rjw6289, txh710\}@psu.edu}  \quad 
\textsuperscript{2}\texttt{changyel@uw.edu}
}

\expandafter\def\expandafter\normalsize\expandafter{%
    \normalsize%
   \setlength{\textfloatsep}{3pt plus 1.0pt minus 2.0pt} 
\setlength{\floatsep}{5pt plus 1.0pt minus 2.0pt}     
\setlength{\intextsep}{5pt plus 1.0pt minus 2.0pt}
}

\usepackage{comment}

\title{``Newspaper Eat'' Means ``Not Tasty'': A Taxonomy and Benchmark for Coded Language in Real-World Chinese Online Reviews}

\newcommand{\eg}{{\it e.g.}\xspace}

\newcommand{\dataset}{\mbox{\textsc{CodedLang}}\xspace}

\newcommand{\changye}[1]{\textcolor{violet}{Changye: #1}}

\begin{document}

\maketitle




\begin{abstract}
Coded language is an important part of human communication.
It refers to cases where users intentionally encode meaning so that the surface text differs from the intended meaning and must be decoded to be understood.
Current language models handle coded language poorly.
Progress has been limited by the lack of real-world datasets and clear taxonomies.
This paper introduces \textbf{\dataset}, a dataset of 7,744 Chinese Google Maps reviews, including \textbf{900 reviews with span-level annotations of coded language}.
We developed a seven-class taxonomy that captures common encoding strategies, including phonetic, orthographic, and cross-lingual substitutions.
We benchmarked language models on coded language detection, classification, and review rating prediction.
Results show that even strong models can fail to identify or understand coded language. 
Because many coded expressions rely on pronunciation-based strategies, we further conducted a phonetic analysis of coded and decoded forms. Our code and dataset are publicly available \footnote{https://github.com/Crowd-AI-Lab/CodedLang}.
Together, our results highlight coded language as an important and underexplored challenge for real-world NLP systems.

\end{abstract}

\section{Introduction\label{sec:introduction}}

\begin{figure}
    \centering
    \includegraphics[width=1\linewidth]{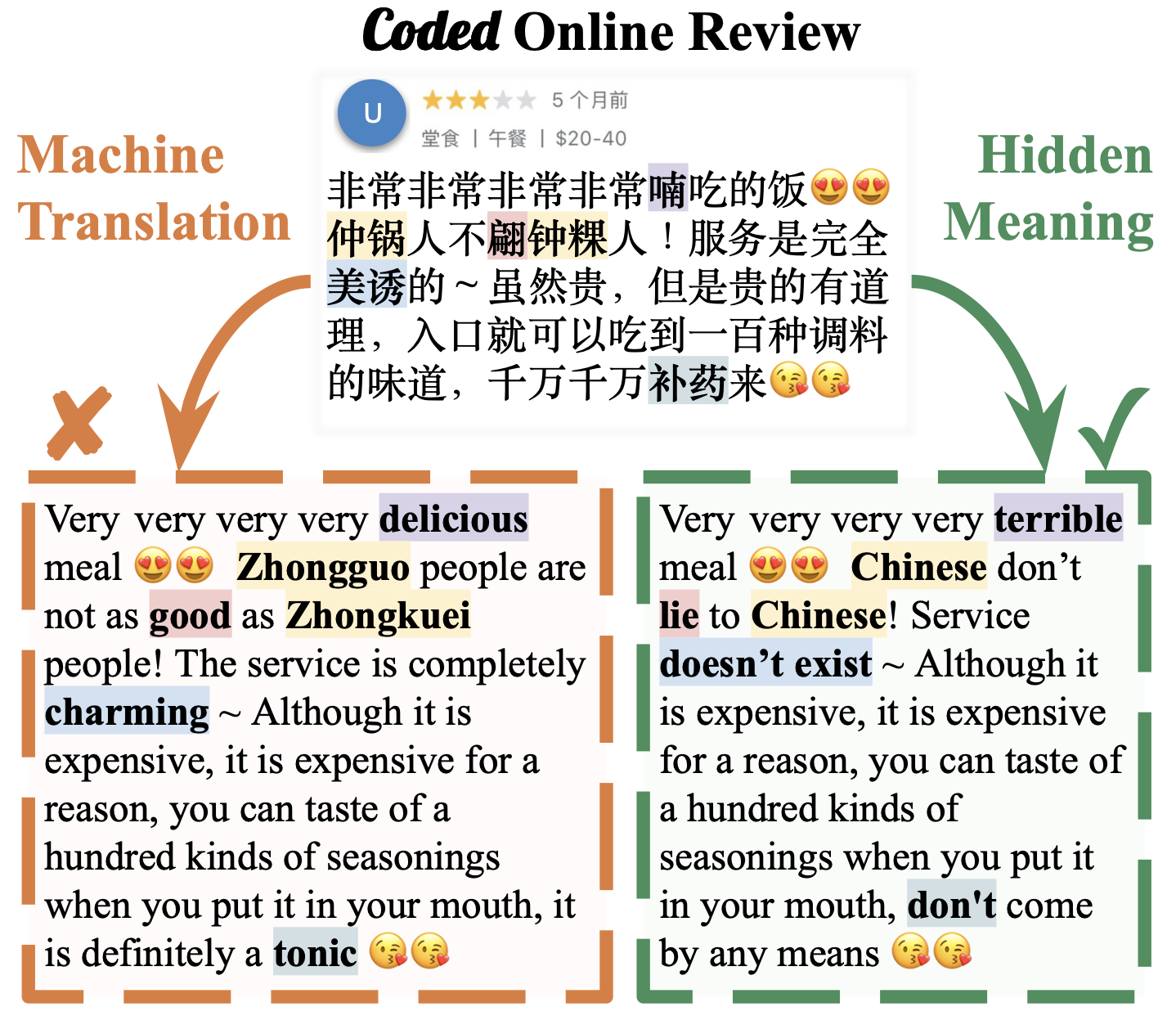}
    \caption{Example of creative coded language in a Google Maps restaurant review. Top: the original user review in Chinese, as posted on Google Maps. Left: the machine-generated English translation produced by Google Translate. Right: the inferred underlying meaning obtained by decoding phonetic substitutions.}
\label{fig:NE}
\end{figure}

Coded language is an important way people communicate with each other~\cite{ji2018creative}.
Instead of expressing meaning directly, writers intentionally encode their intent so that the surface text differs from the intended meaning and must be decoded by an informed reader. 
Such practices are common in everyday communication, yet they remain understudied in natural language processing (NLP), where models typically assume that meaning is conveyed transparently in literal text.

Online review platforms provide a clear real-world context in which coded language emerges.
While reviews influence consumer decisions and public reputation, writing negative reviews can expose users to social pressure, retaliation, or reduced visibility. 
Among Chinese-speaking users on Google Maps, we observed a recurring strategy: embedding negative evaluations inside seemingly positive or neutral reviews using coded language. 
These reviews rely on homophones, phonetic substitutions, transliteration, emoji, and cross-lingual cues that obscure meaning from casual readers, machine translation systems, and moderation tools, while remaining interpretable to in-group audiences.
For example, a review stating ``\textbf{Newspaper eat}'' appears nonsensical in English but encodes a negative message through a phonetic substitution of the Chinese phrase ``\begin{CJK}{UTF8}{gbsn}不好吃\end{CJK}'' (bu4 hao3 chi1, meaning \textbf{not tasty}). 
When spoken quickly, ``bu4 hao3 chi1'' can sound like ``bao4 chi1'' (\begin{CJK}{UTF8}{gbsn}报吃\end{CJK}), which literally translates to ``Newspaper Eat'' in English, where ``bao4'' means newspaper and ``chi1'' means eat.
Figure~\ref{fig:NE} shows another representative example, where machine translation systems fail to recover the intended meaning. 
These examples illustrate how coded language creates a systematic gap between surface form and communicative intent that current NLP systems struggle to bridge.

To study this phenomenon, we curated \textbf{\dataset}, a dataset of 7,744 real-world Chinese Google Maps reviews, including \textbf{900 reviews containing coded language with span-level annotations} (Section~\ref{sec:data}). 
We also developed a seven-class taxonomy that captures common encoding strategies, including homophonic, phonetic, and cross-lingual substitutions (Section~\ref{sec:taxonomy}). 
Using \dataset, we evaluated modern language models on three NLP tasks (Section~\ref{sec:evaluation})
and further analyzed the phonetic properties of coded expressions (Section~\ref{sec:analysis}).
This work treats coded language as a systematic and understudied form of real-world communication and provides a grounded dataset and taxonomy for examining how NLP systems handle intentionally encoded meaning in practice.

\section{Related Work\label{sec:related-work}}
\paragraph{Natural Language Processing Methods for Coded Languages.}
Prior NLP work showed that coded language is hard to analyze due to its dependence on context, cultural knowledge, polysemy, and limited labeled data~\cite{ji2018creative}.
Decoding could be further complicated in multilingual settings, where translation systems and multilingual language models misinterpret coded expressions due to misleading cross-language word similarity~\cite{kallini2025false}.
Open challenges included balancing opacity and interpretability, modeling personalized and multilingual coding strategies, and handling the rapid evolution of code phrases. 
While these challenges were known, prior work lacked task formulations and datasets grounded in real-world review platforms. 
Our work addressed this gap by formulating coded reviews as an NLP task and introducing \dataset, a curated dataset of real-world coded reviews.

Existing studies focused on encoding and decoding mechanisms.
\citet{ji2018creative} showed how users encoded meaning to bypass censorship while remaining intelligible to in-group audiences, supporting functions such as discussing sensitive topics and expressing negative opinions indirectly. 
\citet{kim2021trkic} found that Korean speakers on Google Maps used morphological, phonological, optical, and semantic strategies to make reviews intelligible only to in-group readers.
\citet{cho2021google} also categorized intentionally noisy text by interpretation boundaries.
However, these studies did not capture how users systematically hide negative sentiment inside positive or neutral reviews, nor did they provide span-level annotations for this behavior.
Coded language has also been studied in political discourse as dog whistles, which are expressions that appear innocuous to general audiences but convey specific meanings to targeted in-groups~\cite{mendelsohn2023dogwhistles, xu2021blow}.
This line of work focused on ideological signaling rather than everyday consumer communication and did not consider how coded language interacted with ratings or machine translation.

\paragraph{Phonetic and Orthographic Substitution.}
Early NLP work treated phonetic and orthographic variation as noise to be normalized~\cite{saito2014morphological}.
More recent studies have shown that such variations were deliberately used as coding strategies.
In Chinese internet language, homophonic substitutions and character-level perturbations encode meaning implicitly and require phonological reasoning and shared background knowledge for interpretation~\cite{xu2021blow, ma2025reasoning}.
Benchmarks showed that large language models (LLMs) struggle with these substitutions and often rely on memorized patterns rather than systematic reasoning~\cite{ma2025reasoning, ma2025phonothink}.
Related work in content moderation also showed that homophones, emoji substitutions, and visually similar characters degraded toxic language detection performance~\cite{xiao2024toxicloakcn}.
Recent work also explored countermeasures, showing that simple LLM-based approaches could partially counter algospeak~\cite{fillies2024simple}.
However, this line of work treats surface manipulation as an attack rather than a communication strategy. 
Our work instead studies how real users use phonetic and orthographic substitutions to express negative evaluations while avoiding detection.

\section{\dataset Dataset and Taxonomy\label{sec:data}}
\begin{table*}[t]
\centering
\begin{threeparttable}
\begin{CJK}{UTF8}{gbsn}
\small
\setlength{\tabcolsep}{6pt}
\begin{tabular}{p{3.4cm} p{6cm} p{5.3cm}}
\hline
\textbf{Coded Language Class} & \textbf{Definition} & \textbf{Example} \\
\hline

Ambiguous Homophone &
Existing lexical phrases whose surface form is a valid expression, but whose intended meaning differs from the literal meaning via homophones &
\makecell[l]{``炒鸡''(chao3 ji1, fried chicken) =\\超级 (chao1 ji2, super)} \\

Non-Lexical Homophone &
Homophonic characters that do not form a valid lexical expression on the surface, but can be decoded phonetically into a conventional phrase. &
\makecell[l]{``灰常''(hui1 chang2, grey often) =\\非常 (fei1 chang2, very)} \\

Phonetic Substitution &
Use of phonetic symbols, including Pinyin, Zhuyin, or Roman letters, to substitute for standard Chinese characters based on pronunciation. &
\makecell[l]{
``tm'' = 他妈 (ta1 ma1, damn);\\
``好ㄘ''(hao3 c, good) =\\好吃 (hao3 chi1, tasty)}\\

Emoji Substitution &
Emoji replace textual content and convey an encoded meaning that differs from the literal emoji semantics. &
\makecell[l]{``\includegraphics[height=1.2em]{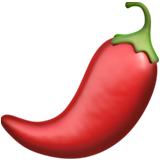}\includegraphics[height=1.2em]{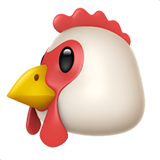}''
=\\辣鸡 (la4 ji1, spicy chicken) =\\垃圾 (la1 ji1, trash)} \\

Orthographic Substitution &
Encoding through visual or orthographic similarity between characters and symbols. &
\makecell[l]{``口乞”(kou3 qi3, mouth begging) =\\吃 (chi1, eat)} \\

\makecell[l]{Cross-Lingual\\Phonetic Encoding} &
Encoding meaning through cross-lingual phonetic approximation, often involving transliteration or translation of homophones. &
\makecell[l]{``古德(gu3 de2, old virtue)'' = Good;\\
``Newspaper eat'' =\\报吃 (bao4 chi1, newspaper eat) =\\不好吃(bu4 hao3 chi1, not tasty)
}\\

Cipher &
The whole sentence is a nonsensical expression without a clear literal semantic meaning. &
``qqQQaQ！你燃yfyi y f r'' \\

\hline
\end{tabular}
\end{CJK}
\caption{Taxonomy of coded language classes with definitions and representative examples. Examples are presented as they appear in the original reviews, with the decoded form, Pinyin, and English meanings provided in parentheses. Cipher instances are not decoded, as they are not directly interpretable.}
\label{tab:taxonomy}
\end{threeparttable}
\end{table*}


We curated \dataset, a dataset of 7,744 online reviews, of which 900 contain coded language. 
Each coded-language review includes span-level annotation(s) indicating the coding strategy, as defined by our taxonomy.
This section overviews the construction and analysis of \dataset.


We defined \textbf{coded language} as \textit{the intentional encoding or obfuscation of meaning, in which the surface form differs from the intended meaning and decoding is required for interpretation}.
Importantly, not all socially opaque 
language qualifies as coded language. 
Dialect, jargon, and multilingual text can be socially opaque because they rely on shared community knowledge, but they are not considered coded language unless the meaning is intentionally disguised through an explicit encoding mechanism, such as phonetic, orthographic, or symbolic substitutions. 
In contrast, coded language is socially opaque because it is encoded: understanding it requires actively decoding the surface form to recover the intended meaning.


\subsection{Data Sources and Pre-Processing\label{sec:data-source}}

\paragraph{Initial Coded-Language Examples.}
We first observed news articles reporting Chinese-speaking users employing coded language to express dissatisfaction in Google Reviews for overseas businesses~\cite{shine2024newspaper, radii2024coded, yahoo_news_2025}. 
These cases were originally shared on social media platforms such as Xiaohongshu (Rednote) and Threads, where users discussed the opacity and in-group interpretability of such reviews and shared additional examples they had encountered.\footnote{For example, a Xiaohongshu post titled ``This is the first time I've seen real Chinese encryption hahaha'', and a Threads post said ``Saw a review that only Taiwanese people can understand, thank you, kind soul!'', both highlighted community awareness of such coded expressions.}
Therefore, we collected 23 reviews containing coded language that were shared on social media or reported in news articles. 
These reviews serve as an initial seed set, providing concrete examples of real-world coded expressions and guiding our large-scale mining of low-frequency coded reviews. 

\paragraph{Real-World Google Maps Reviews.}
Building on this seed set, we further screened for coded-language instances in massive Google Maps reviews. 
We used two Google Maps review datasets:
{\em (i)} 1.77 million restaurant reviews~\cite{li2022uctopic} and 
{\em (ii)} 666 million local reviews~\cite{yan2023personalized}.
These reviews were multilingual and covered U.S. businesses, including restaurants and hotels.
Each review included a user ID, a business ID, the original review text, ratings, and an English translation of the review when the original was not in English.
We then selected reviews containing at least one Chinese character,
which yielded 5,391 Chinese-language reviews from the restaurant review dataset and 112,521 from the local review dataset. 
This set included reviews written in Simplified and Traditional Chinese, with some mixed with English and other languages.\footnote{Here, \emph{Traditional Chinese} refers to written Chinese in traditional characters, including varieties such as Taiwan Mandarin and Cantonese.}
Section~\ref{sec:data-mining} described our process of finding reviews that contain coded languages from this set.


\subsection{Constructing \dataset Dataset\label{sec:data-mining}}


We used an iterative, human-in-the-loop mining process to identify coded languages in large-scale Google Maps reviews.
Two of the authors served as annotators in this process.



\paragraph{Pilot Study.}
We collected 23 gold examples of coded language from social media and news media discussions (sources described in Section~\ref{sec:data-source}).
To estimate real-world prevalence, we manually annotated 5,391 U.S. restaurant reviews (created in Section~\ref{sec:data-source}). 
We identified only 19 reviews (0.35\%) that used coded language, confirming that it is a rare phenomenon in this domain.
We also found that most coded expressions occur at the \textit{phrase level} rather than the sentence level, meaning they can often be decoded using limited local context.
A dictionary-based table lookup covering common phrases, combined with substantial human validation, should capture a large portion of real-world cases.




\paragraph{Step 1: Seed Dictionary Construction.}
We first annotated the specific text spans that contained coded language in all examples (\eg,``\begin{CJK}{UTF8}{gbsn}虾\end{CJK}'' in ``\begin{CJK}{UTF8}{gbsn}真的是虾到爆\end{CJK}'') from Pilot Study, where \begin{CJK}{UTF8}{gbsn}虾\end{CJK} (xia1, shrimp) is a homophonic substitution for \begin{CJK}{UTF8}{gbsn}瞎\end{CJK} (xia1, absurd), and then aggregated these annotated spans to form a dictionary.
The initial dictionary contained 100 coded spans. 

\paragraph{Step 2: Candidate Retrieval and Validation.}
Using this dictionary, we applied string matching across the entire review corpus to retrieve candidate reviews containing dictionary entries. 
To further expand coverage, we also included reviews whose English machine translations contained Pinyin, a phonetic system for Chinese that uses the Latin alphabet to represent pronunciation, since this indicates potential machine translation challenges. 
All candidate reviews were then manually validated. 


\paragraph{Step 3: Dictionary Expansion and Iterative Bootstraping.}
The validated reviews from Step 2 went through the same span-annotation process described in Step 1, and the newly annotated spans were added to the dictionary. 
The updated dictionary was then used to retrieve additional coded-language reviews from the corpus, and this process was repeated iteratively.
We continued the bootstrapping process until convergence, defined as an iteration in which no new validated coded-language spans were identified and no additional coded-language reviews were retrieved.
To further improve coverage, we also incorporated all word entries from an external homophone dictionary derived from Weibo and Tieba data~\cite{ma2025reasoning} during this iterative process.

As a result, \dataset contained \textbf{900 reviews containing coded language} out of 7,744 manually annotated reviews.
These 7,744 items include the 5,391 annotated restaurant reviews from the Pilot Study, as well as cases surfaced throughout the iterations.

\paragraph{Annotating Coded Language Classes.}
Based on the accumulated coded-language instances, we developed a taxonomy of seven coded-language categories (Section~\ref{sec:taxonomy}) and annotated each review using multi-label class assignments. 
During annotation, annotators first identified whether a review contained coded language, then marked the exact text spans that were coded. 
Each coded span, consisting of one or more characters, was assigned a single category corresponding to its primary encoding strategy.  
It is possible for a single span to involve multiple encoding mechanisms. But for annotation consistency and downstream evaluation clarity, we adopted a one-to-one span–category mapping, assigning each span to its dominant encoding strategy. Our annotation decision was based on both surface form and intended meaning. While some expressions may involve layered transformations, we categorize them according to the most salient observable mechanism. For example, the expression `` 
\begin{CJK}{UTF8}{gbsn}铁\end{CJK}\includegraphics[height=1.2em]{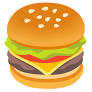}\includegraphics[height=1.2em]{latex/figures/hamburger.png}'' derives from the phrase ``\begin{CJK}{UTF8}{gbsn}铁憨憨\end{CJK}'' (tie3 han1 han1, meaning ``silly''). The transformation involves an intermediate phonetic substitution \begin{CJK}{UTF8}{gbsn}憨\end{CJK} (han1, silly) to \begin{CJK}{UTF8}{gbsn}汉\end{CJK} (han4), to \begin{CJK}{UTF8}{gbsn}汉堡\end{CJK} (han4 bao3, hamburger) which is  \includegraphics[height=1.2em]{latex/figures/hamburger.png}, but the final surface form primarily manifests as emoji substitution. Therefore, it is labeled under emoji substitution in our taxonomy. 
A review could contain multiple spans using different strategies and thus receive multiple coded-language labels. This annotation process resulted in 184 unique coded spans across the dataset. While each span was assigned a single encoding category, its decoded meaning could vary depending on contextual usage.

\paragraph{Inner-Annotator Agreements.}
During the iterative bootstrapping phase, the first author manually labeled all items returned by the process and included all labeled items, both coded and non-coded, in \dataset. 
The second author then independently annotated the entire \dataset. 
They achieved high agreement on whether a review contained coded language (Cohen's kappa = 0.99), which was calculated on the annotated data before resolving disagreements.. At the span level, annotators also showed strong consistency, with 98\% of identified spans overlapping between annotators. Recurring disagreements primarily arose in distinguishing dialectal variation from intentionally coded language. In some cases, regional phonetic spellings resembled coded substitutions, requiring annotators to determine whether a form reflected deliberate encoding or ordinary dialectal expression. For example, \begin{CJK}{UTF8}{gbsn}里格楞\end{CJK} was a candidate of coded expression, but was later identified as a Beijing dialect term meaning `pull strings'. We labeled it as non-coded language, as its surface form directly reflects its intended meaning, despite posing an interpretation barrier for those unfamiliar with the dialect.
All disagreements were resolved through discussion to reach consensus, which determined the final released labels.

\subsection{Taxonomy of Coded Language\label{sec:taxonomy}}

\begin{figure}
    \centering
    \includegraphics[width=1\linewidth]{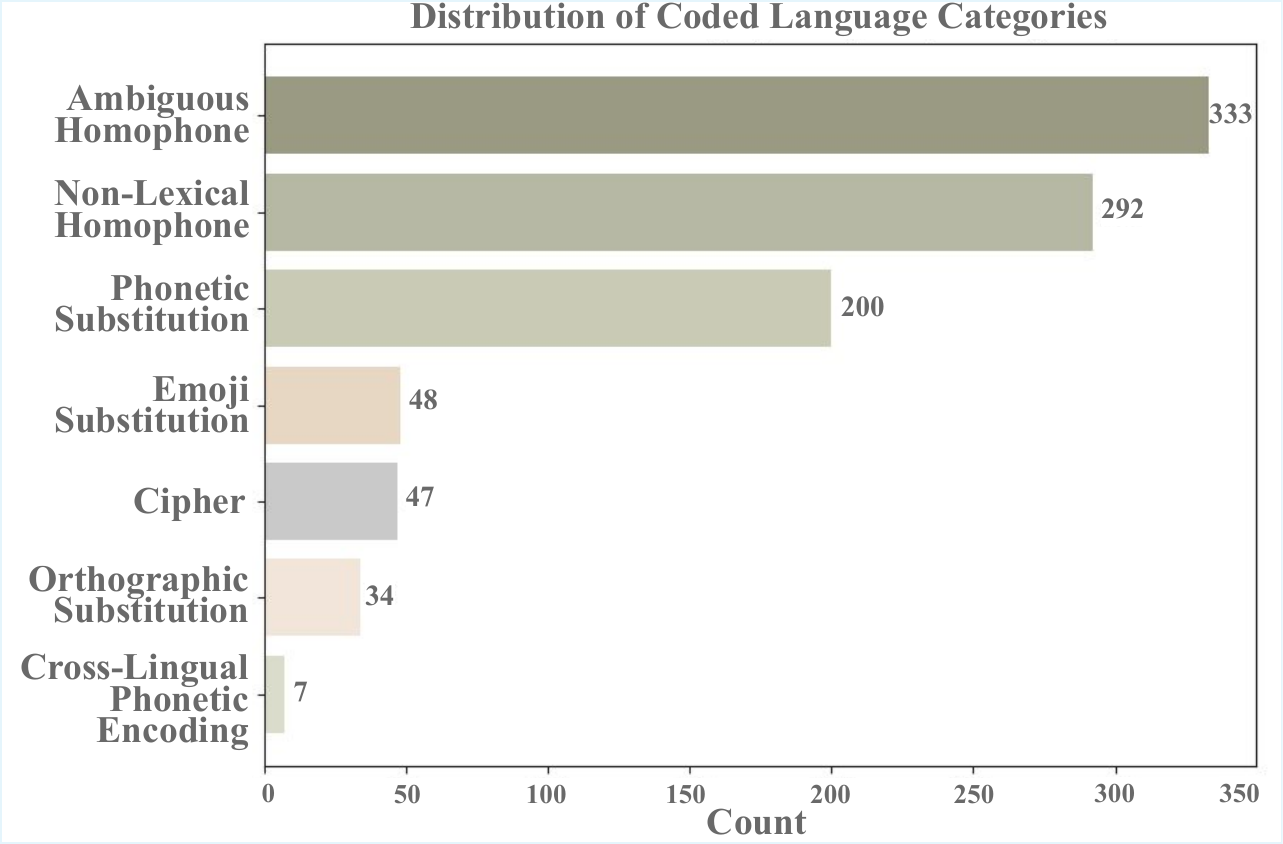}
    \caption{Distribution of coded language categories in the annotated dataset.}
\label{fig:category_dist}
\end{figure}

We inductively refined the decision boundaries of each category through repeated discussions between the two annotators through multiple rounds of review. 
Table~\ref{tab:taxonomy} summarizes the final, converged taxonomy, including concise definitions and representative examples.
Figure~\ref{fig:category_dist} shows the distribution of coded-language categories. Homophone-based strategies are the most prevalent, followed by phonetic and orthographic substitutions, while cipher-like expressions and cross-lingual phonetic encoding are rarer.

\section{Downstream Benchmark Tasks \label{sec:evaluation}}
Using \dataset, we evaluated the capability of LLMs to handle real-world coded language across three NLP tasks: 
{\em (i)} coded language detection, 
{\em (ii)} coded language classification, and 
{\em (iii)} review rating prediction.

\subsection{Coded Language Detection\label{sec:detection-task}}

\paragraph{Experimental Setups.}
We assessed LLMs' ability to detect coded language in reviews using a binary classification task. 
We benchmarked a diverse set of modern English and Chinese LLMs, including GPT-5-mini~\cite{openai2025gpt5mini}, Gemini-2.5-Flash~\cite{comanici2025gemini25pushingfrontier}, Qwen2.5-7B-Instruct~\cite{qwen2025qwen25technicalreport}, DeepSeek-V3.2~\cite{deepseekai2025deepseekv32pushingfrontieropen} and Llama-3.1-8B-Instruct~\cite{llama3_1}.
Models were instructed to identify whether a review contains coded language and output a binary prediction. 
Our prompt included the definition of coded language along with few-shot examples drawn from our taxonomy.
We evaluated classification performance using precision, recall, and F$_1$ score.

\paragraph{Results.}

\begin{table}[t]
\centering
\small
\begin{tabular}{lccc}
\toprule
\textbf{Model} & \textbf{Precision} $\uparrow$ & \textbf{Recall} $\uparrow$ & \textbf{F$_1$} $\uparrow$ \\
\midrule
GPT-5-mini & 0.69 &	0.80 & 0.74  \\
Gemini-2.5-Flash & 0.67 & \textbf{0.92} & 0.77 \\
DeepSeek-V3.2   & 0.77 & 0.79 & \textbf{0.78} \\
Qwen2.5-7B-Instruct & \textbf{0.80} & 0.43 & 0.56 \\
Llama-3.1-8B-Instruct & 0.35 & 0.68 & 0.46 \\
\bottomrule
\end{tabular}
\caption{Coded language detection performance across models. Best scores in each column are in \textbf{bold}.}
\label{tab:detection_results}
\end{table}

From the Table \ref{tab:detection_results}, we found DeepSeek-V3.2 has a balanced tradeoff between precision and recall, and has the highest F$_1$ score in the coded language detection. Gemini-2.5-Flash has the highest recall, which captures the majority of coded expressions at the cost of reduced precision. On the other side, Qwen2.5-7B-Instruction has the highest precision.
Comparing across four models, there is no single model that dominates across all metrics.

\begin{table}[t]
\centering
\small
\setlength{\tabcolsep}{3pt}
\begin{tabular}{lcccc}
\toprule
\textbf{Coded Language Class} &
\textbf{P} $\uparrow$ &
\textbf{R} $\uparrow$ &
\textbf{F$_1$} $\uparrow$ &
\textbf{N}\\
\midrule
Ambiguous Homophone     & 0.48&	\underline{0.46} &	0.47 &333 \\
{Non-Lexical Homophone}  &0.62&	0.57&	0.59& 292\\
Phonetic Substitution    & 0.47&	0.56&	0.51 & 200\\
Emoji Substitution       & 0.36&	0.50&	0.42& 48\\
Cipher & \textbf{0.84}&	0.66&	\textbf{0.74} &47 \\
{Orthographic Substitution} &0.32&	0.94&	0.48 & 34 \\
{Cross-Lingual 
Phonetic Encoding} & \underline{0.25}&	\textbf{1.00}&	\underline{0.40} & 7\\
\bottomrule
\end{tabular}
\caption{Multi-label coded language classification performance across categories by \texttt{DeepSeek-V3.2}. The highest (lowest) scores in each column are highlighted in \textbf{bold} (\underline{underlined}).}
\label{tab:classification_by_class}
\end{table}

\subsection{Coded Language Classification\label{sec:classification-task}}

\paragraph{Experimental Setups.}
We formulated coded language classification as a multi-label classification task, where a review may apply multiple coded language strategies to encode its message. Our prompts provide concise definitions of each coded-language category along with a few-shot representative examples. Given a review, models are instructed to produce a structured output with three fields: (1) classes, a list of coded-language categories applied in the review; (2) evidence, a list of exact text spans copied from the review that support the classification; and (3) a brief explanation justifying the decision. The output classes and the evidence of the coded span would be empty if no coded language is detected. Models are evaluated against annotations derived from our taxonomy and measured using precision, recall, and F$_1$ score.

\paragraph{Results.}
We reported the results of the coded-language classification using DeepSeek-V3.2. Results for other models are provided in the Appendix \ref{sec:additional_classification}. Table \ref{tab:classification_by_class} reports multi-label classification performance across coded language categories. To account for class imbalance, we reported per-class performance metrics, allowing a granular assessment across categories. Cipher expressions achieved the highest F$_1$ and precision, indicating that they are easier to identify than other categories. Cross-Lingual Phonetic Encoding had the highest recall but lowest precision and F$_1$, which tells that the model can not identify this category clearly. This might be influenced by the limited sample size of the Cross-Lingual Phonetic Encoding Category. The Homophones, Phonetic, Orthographic, and Emoji Substitutions achieved F$_1$ scores ranging from 0.42 to 0.59, indicating that these phenomena remained challenging for current models.

\subsection{Review Rate Prediction\label{sec:rating-prediction-task}}
\begin{table*}[t]
\centering
\small
\begin{tabular}{lccccc}
\toprule
\textbf{Model} &
\textbf{Non-coded} $\downarrow$ &
\textbf{Decoded} $\downarrow$ &
\textbf{Span-Masked} $\downarrow$ &
\textbf{Coded} $\downarrow$ &
\textbf{Char-Masked} $\downarrow$ \\
\midrule
GPT-5-mini    & \textbf{0.52} & 0.55 & 0.65 & 0.68 & \underline{0.71} \\
Gemini-2.5-Flash   & \textbf{0.63} & 0.69 & 0.73 & 0.70 & \underline{0.74} \\
DeepSeek-V3.2 &  \textbf{0.47}& 0.82 & \underline{1.16} & 0.80 & 1.08 \\
Qwen2.5-7B-Instruct  & \textbf{0.90} & 1.27 & 1.60 & \underline{1.90} & 1.67 \\
Llama-3.1-8B-Instruct & \textbf{0.67} & 0.89 & 0.93 & \underline{1.30} & 1.23 \\
\bottomrule
\end{tabular}
\caption{Review rating prediction performance (mean squared error (MSE), lower is better) across different review types. The best performance in each row is highlighted in \textbf{bold}, and the worst is \underline{underlined}. Overall, LLMs perform worse when processing reviews that contain coded language.}
\label{tab:rating_pred_mse}
\end{table*}

\begin{figure}
    \centering
    \includegraphics[width=1\linewidth]{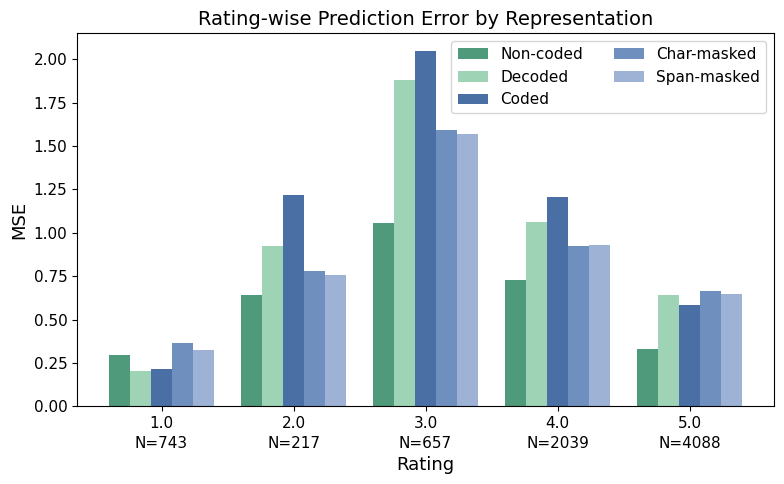}
    \caption{Rating-wise mean squared error (MSE) for review rating prediction under different textual representations. Results were predicted by \texttt{GPT-5-mini}. Sample sizes for each rating level were shown below the x-axis, reflecting imbalance across ratings.}
    \label{fig:rating_pred}
\end{figure}

\paragraph{Experimental Setups.}
We examine review rate prediction to understand how coded language affects models' ability to infer users' intended star ratings. While ratings are commonly treated as proxies for sentiment, reviews that use coded language may express evaluative intent indirectly, introducing additional ambiguity. Considering the misalignment between ratings and review content, we manually validated the pairs of ratings and reviews. We found that ratings are generally consistent with review sentiment, with only seven reviews rated four or five that express negative content. 

The review rating prediction was evaluated on Mean Squared Error (MSE). To investigate the effect of coded language, we construct multiple formats of the same review text: (1) Non-coded reviews as original; (2) Coded reviews as original; (3) Character-based masked reviews, where each coded characters is replaced with a [CODE] token; (4) Span-based masked reviews, where consecutive coded tokens are replaced with a single [CODE-SPAN]; and (5) Decoded reviews, where coded expressions are replaced with their intended meaning using a dictionary derived from annotation. 

\paragraph{Results.}
Across all models in Table ~\ref{tab:rating_pred_mse}, the MSE on coded reviews is larger than on non-coded reviews, indicating higher prediction error under coded language. It shows that coded language disrupted sentiment signals relevant for rating prediction. GPT-5-mini showed the most robust performance across review formats, with consistently low MSE, so we further examined its predictions stratified by rating groups.

Figure~\ref{fig:rating_pred} presents rating-wise prediction error across different textual representations of reviews. Prediction errors were low for extreme ratings (1 and 5), indicating that strongly expressed sentiment was relatively easy to infer even in the presence of coded or masked language. In contrast, 3-star reviews exhibited the highest error, reflecting the inherent ambiguity of mixed or neutral evaluative intent. Notably, for intermediate ratings (2-4), masked reviews have lower error than decoded and coded reviews, suggesting that coded expressions can introduce misleading surface cues and that literal decoding does not always clarify intent.


As an additional analysis, we found that
coded language was significantly associated with lower ratings in Google Reviews ($p<0.001$). 
Coded language tended to appear in relatively concise reviews under extreme sentiment conditions, even though negative reviews were longer overall ($p<0.001$). 
Appendix~\ref{sec:stat} shows the detailed analysis.


\section{Phonetics Analysis\label{sec:analysis}}


LLMs often performed poorly on coded language, but it remained unclear which factors drive this performance drop. 
In \dataset, many coded reviews relied on pronunciation-based strategies, which raised an interesting question: 
\textbf{Does phonetic similarity between coded expressions and their intended meanings make coded language easier for LLMs to process?}
We conducted a set of analyses to answer this question.

\begin{table}[t]
\centering
\small
\begin{tabular}{lcc}
\toprule
\textbf{Coded Language Class} & \textbf{CER-Pinyin} & \textbf{CER-IPA} \\
\midrule
Ambiguous Homophone & 0.49 & 0.57 \\
Non-Lexical Homophone & \textbf{0.30}& \textbf{0.37}\\
Phonetic Substitution & 0.61 & 0.68 \\
Emoji Substitution & 0.58 & 0.63 \\
\makecell[l]{Cross-Lingual\\Phonetic Encoding}& \underline{1.23} & \underline{2.46} \\
\bottomrule
\end{tabular}
\caption{Character error rates (CER) between coded expressions and their decoded forms. Lower CER indicates greater pronunciation similarity. The lowest (highest) CER are highlighted in \textbf{bold} (\underline{underlined}).}
\label{tab:cer_pinyin_ipa}
\end{table}

\begin{table}[t]
\centering
\small
\begin{tabular}{l c}
\toprule
\textbf{Coded Language Class} & \textbf{MSE} \\
\midrule
Ambiguous Homophone & 0.31 \\
Non-Lexical Homophone & 0.43 \\
Phonetic Substitution & 0.51 \\
Emoji Substitution & 0.56 \\
Cross-Lingual Phonetic Encoding & 0.86 \\
\bottomrule
\end{tabular}
\caption{Mean squared error (MSE) between GPT5-mini  predicted and user-assigned ratings across coded language classes.}
\label{tab:mse_by_class}
\end{table}

\begin{figure}[t]
\centering
\includegraphics[width=1\linewidth]{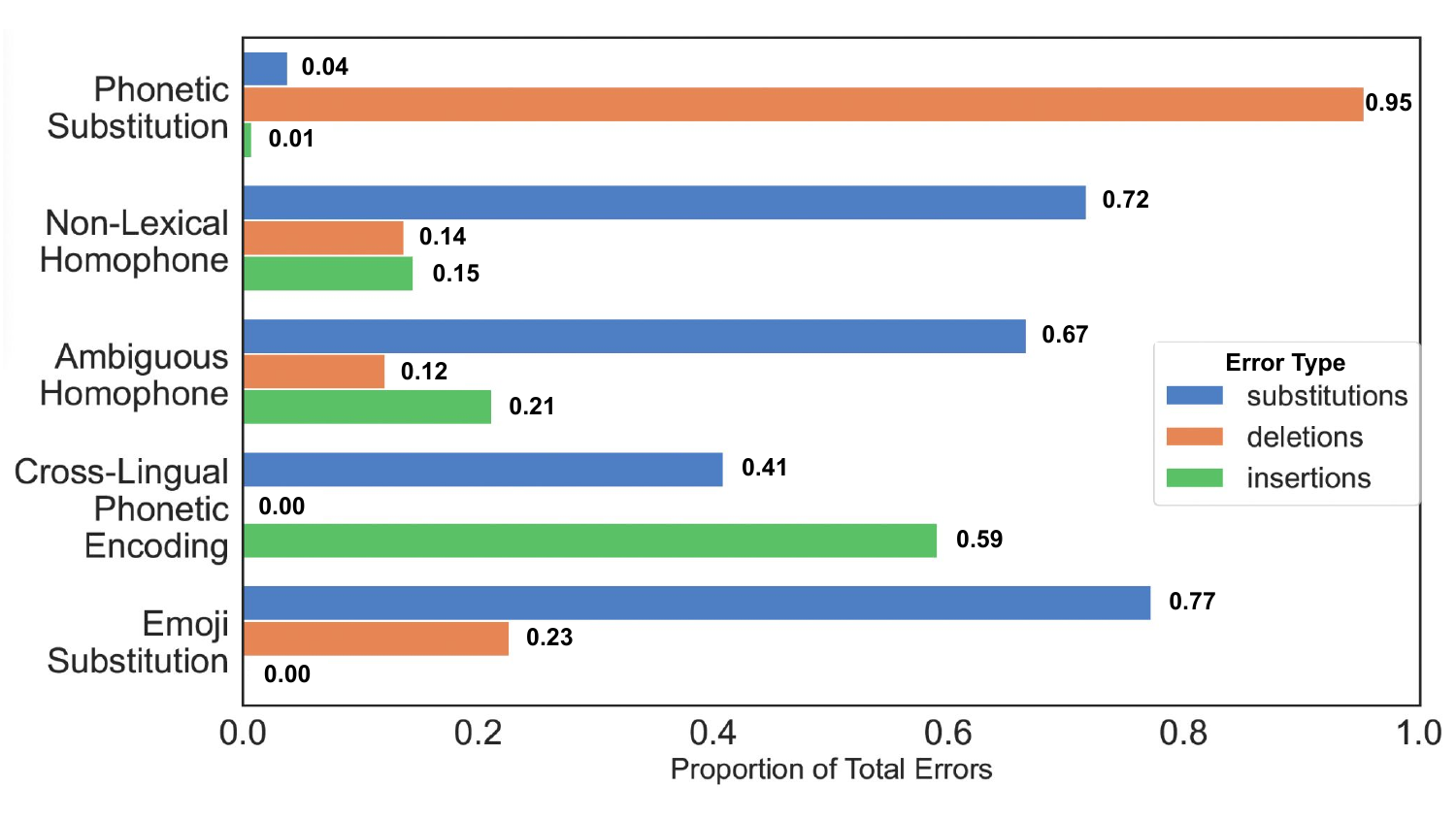}
\caption{The average error rate for substitutions, deletions, and insertions per class using Pinyin representations, ranked by the total number of errors per class.}
\label{fig:pinyin}
\end{figure}

\begin{figure}[h]
\centering
\includegraphics[width=1\linewidth]{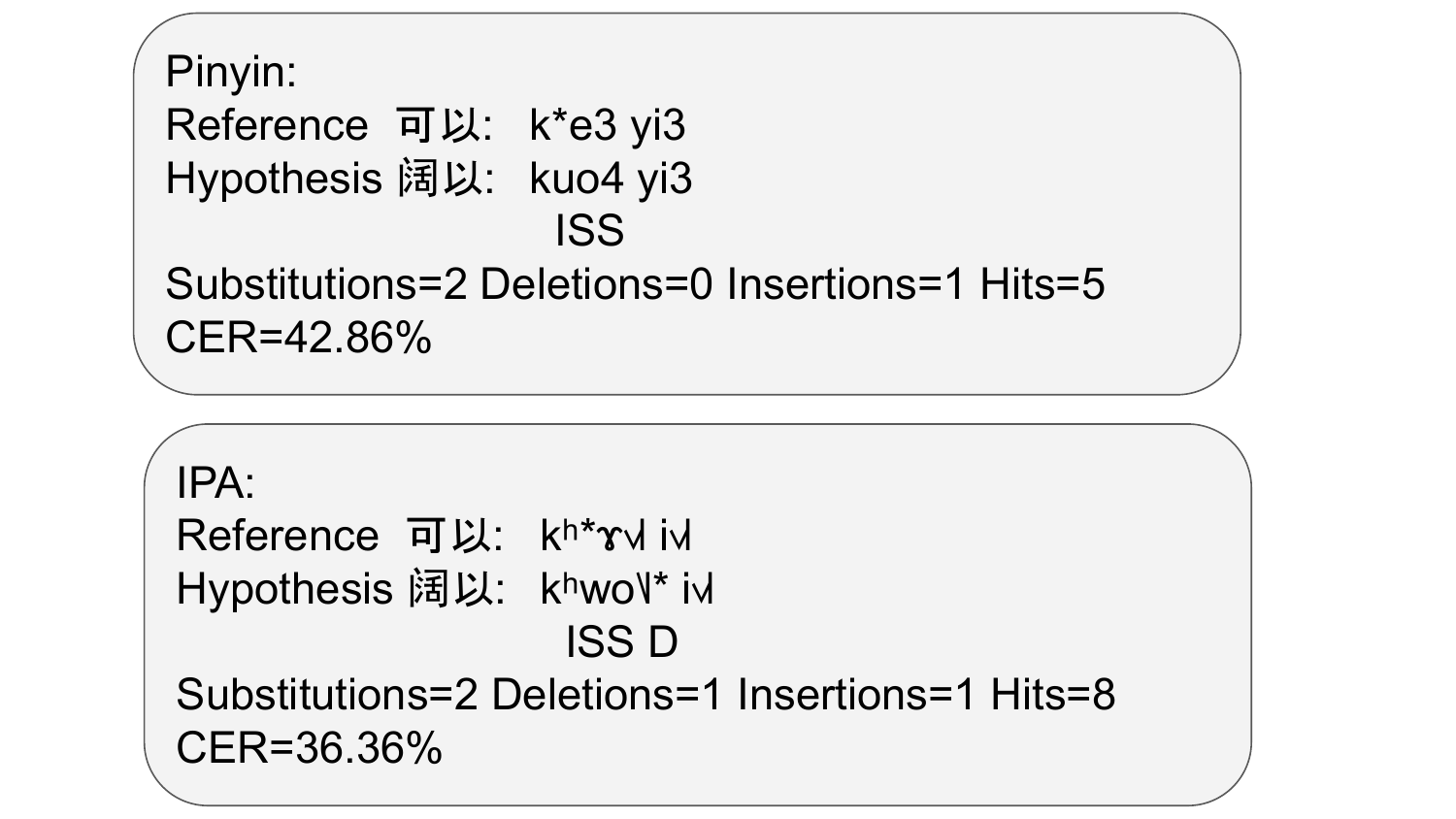}
\caption{
Example of CER based on Pinyin and IPA representations.
}
\label{fig:CER_example}
\end{figure}

\paragraph{Measuring Pronunciation Similarity.}
We measured phonetic similarity between coded expressions and their decoded meanings. 
We used tools to annotate Pinyin~\cite{huang2025pypinyin} and International Phonetic Alphabet (IPA) representations~\cite{taubert2025pinyintoipa} for each coded-decoded span pair in \dataset, excluding the Cipher and Orthographic Substitution classes.
Tones in Pinyin were annotated from 1 to 4 and mapped to IPA tone contours accordingly (1 to \tone{55}, 2 to \tone{35}, 3 to \tone{313}, and 4 to \tone{51}). 
We used character error rate (CER), a common evaluation metric in automatic speech recognition (ASR), to quantify the phonetic similarity between coded and decoded spans. In ASR, CER is computed as the normalized sum of 
substitutions ($S$), 
deletions ($D$), and 
insertions ($I$) over the total number of characters ($N$) between the ASR-generated utterance (hypothesis) and verbatim utterance (reference)~\citep{inproceedings} (Equation~\ref{eq:cer}).
In our analysis, a lower CER indicates greater pronunciation similarity between the coded expression and its decoded form.
We treated each phoneme in IPA and syllable in Pinyin as a character, and computed pairwise CER for each pair of decoded span (reference) and coded span (hypothesis). Some pinyin abbreviations consist only of consonant initials, which cannot be directly mapped to IPA without a vowel. In such cases, we follow the common convention used by Chinese speakers when reading pinyin initials aloud, using pronunciation for each alphabet letter without tone to ensure pronounceability before converting to IPA. This avoids generating transduction error [UNK] tokens during IPA conversion, while we acknowledge that this transformation may introduce higher IPA error rate. 
\begin{equation}
\text{CER} = \frac{S+D+I}{N}
\label{eq:cer}
\end{equation}
Figure~\ref{fig:CER_example} illustrates how CER is computed using Pinyin and IPA representations for the same coded expression and its decoded form, including substitution, deletion, and insertion errors.

\subsection{Findings}

\paragraph{Phonetic Error Patterns Across Categories.}

As shown in Table~\ref{tab:cer_pinyin_ipa}, cross-lingual phonetic encoding yielded the highest CER, highlighting the difficulty of cross-language pronunciation mapping. Non-lexical homophones showed the lowest CER, reflecting relatively high phonetic similarity despite semantic ambiguity. 
Figure~\ref{fig:pinyin} shows the average substitution, deletion, and insertion error rates across coded language categories computed using Pinyin, following Equation~\ref{eq:error_rate}. Corresponding results using IPA are reported in Appendix~\ref{app:IPA}.

\begin{equation}
    \text{Error Rate} = \frac{\text{total \# of S/I/D}}{\text{total \# of S+I+D}}
    \label{eq:error_rate}
\end{equation}

We observed that the Phonetic Substitution class reached the highest number of deletions for both Pinyin and IPA. This observation was primarily driven by the use of Pinyin initials, which reduced multi-syllabic words to only Pinyin consonants and subsequently led to failures in the IPA conversion. For example, ``\texttt{rmb}'' was abbreviated from \texttt{\textbf{r}en2 \textbf{m}in2 \textbf{b}i4} (Chinese Yuan).
We also observed a cross-lingual case where ``\texttt{ww}'' was coded for \begin{CJK}{UTF8}{gbsn}哈哈\end{CJK} (haha, laughter), similar to \texttt{lol} (lots of laughs). The origin of \texttt{w} was abbreviated from \begin{CJK}{UTF8}{min}わら\end{CJK} (\texttt{warai}, laughter) in Japanese. In addition, substitutions reached the second most frequent error type across all coded language categories. For example, \texttt{\textbf{ji}n1 ho\textbf{u4} \textbf{xia2}} (\begin{CJK}{UTF8}{gbsn}金厚呷\end{CJK}, non-lexical homophones for `really delicious' in Taiwanese) was substituted for \texttt{\textbf{zhe}n1 h\textbf{a}o\textbf{3} \textbf{chi1}} (\begin{CJK}{UTF8}{gbsn}真好吃\end{CJK}, really delicious). 
This suggested a high degree of phonetic interference, in which the target sounds were replaced by phonetically similar but intentionally incorrect variants.

\paragraph{Phonetic Similarity and Rating Prediction Error.}
We found that greater phonetic distortion in coded language was associated with higher rating prediction error, but this relationship varied across coded language categories and remained modest overall. 
The MSE between the GPT-5-mini predicted rating and the user-assigned rating are shown in the table \ref{tab:mse_by_class}.
The Spearman's $\rho$ between MSE and average CER reached 0.80 for both Pinyin and IPA, suggesting that more character errors were associated with more variance between predicted ratings and user-assigned ratings. Yet, the correlations did not reach conventional levels of statistical significance (p-value = 0.104).

The Cross-Lingual Phonetic Encoding class achieved the highest MSE and CER for both Pinyin and IPA, which may be due to the lack of cross-lingual context and the limited data size. 
Compared to Ambiguous Homophones, Non-Lexical Homophones achieved lower CER under both Pinyin and IPA, but obtained higher MSE in rating prediction. This suggested that, despite higher pronunciation similarity, the loss of lexicalized meaning in Non-Lexical Homophones impairs downstream pragmatic interpretation more.

\section{Discussion\label{sec:discussion}}
\paragraph{Expressiveness vs. Interpretability in Coded Language.}
Our data shows a highly skewed distribution in coded language usage: 
among 184 unique coded spans, 111 appear only once, while a small number of forms are used repeatedly. 
The most frequent span, ``\begin{CJK}{UTF8}{gbsn}滴\end{CJK}''(di1, drip), appears 102 times and serves as a phonetic substitution for ``\begin{CJK}{UTF8}{gbsn}的\end{CJK}''(de, of), often used to convey a playful, cute tone. 
This pattern suggests that while the majority of coded expressions are context-specific, a small subset becomes widely shared and stabilized through repeated use. 
This observation aligns with prior work showing that popular code words rely on unconventional character combinations and simple, memorable, affective, or sarcastic expressions~\cite{zhang2014appropriate}.
At the other extreme, some unique spans in our dataset were difficult even for human annotators to interpret. 
For example ``\includegraphics[height=1.2em]{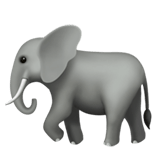}\begin{CJK}{UTF8}{gbsn}人\end{CJK}\includegraphics[height=1.2em]{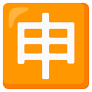}\begin{CJK}{UTF8}{gbsn}一\end{CJK}\includegraphics[height=1.2em]{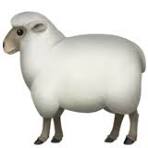}\includegraphics[height=1.2em]{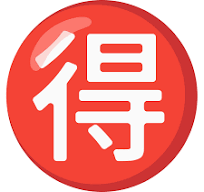}\begin{CJK}{UTF8}{gbsn}姜\end{CJK}'' used emojis to replace homophone words. 
The review reads as (xiang4 ren2 shen1 yi\textbf{1} yang\textbf{2} de\textbf{2} jiang1, elephant human apply one sheep get ginger), which actually means (xiang4 ren2 shen1 yi\textbf{2} yang\textbf{4} de jiang1, the ginger was like ginseng).

There is a trade-off between expressiveness and interpretability in coded language. 
Codes that are easy to recognize, humorous, or representative are more likely to be widely adopted and circulate as shared memes, but their obfuscatory or boundary-setting function weakens as they spread. 
In contrast, niche or opaque codes preserve stronger hiding effects but require greater interpretive effort and shared background knowledge. 
Prior work shows that human-created code words are perceived as funnier than machine-generated ones and are therefore more likely to be adapted over time~\cite{zhang2015context}. 
Similarly, \citet{wan2025hashtag} shows that boundary-making strategies based on linguistic tricks lose effectiveness as audiences diversify, forcing users to continually invent new, more creative, and less transparent codes. 


\paragraph{Multilingualism of Coded Language.} 
Our analysis further shows that coded language often operates across languages and writing systems, creating barriers to interpretation for both native speakers and language models.
Some coded expressions exploited cross-script visual similarity rather than phonetic or semantic equivalence, such as using Kanji or Chinese characters whose shapes resemble Latin letters (\eg, ``\begin{CJK}{UTF8}{gbsn}爪丫\end{CJK}'' encoding ``MY'').
More broadly, Chinese comprises a family of languages and dialects with varying alignments between written and spoken forms.
For example, written Cantonese closely reflects spoken Cantonese and differs substantially from Mandarin in syntax and lexicon, introducing additional variation in coded expressions.
Chinese online communication also draws on sustained contact with neighboring languages such as Japanese and Korean, leading to the use of transliteration-based codes.
For example, ``\begin{CJK}{UTF8}{gbsn}卡哇伊\end{CJK}'' (ka3 wa1 yi1, card wow they) means \begin{CJK}{UTF8}{min}かわいい\end{CJK} (Kawaii, cute) from Japanese. 
These multilingual and cross-script practices highlight that understanding coded language often requires shared community knowledge spanning multiple linguistic systems, further complicating both human annotation and automated interpretation.

\section{Conclusion\label{sec:conclusion}}
Coded language is a common communicative strategy in which intended meaning diverges from surface form and requires decoding to be understood. We presented \dataset, a dataset of 7,744 Chinese Google Maps reviews with span-level annotations and a seven-class taxonomy capturing phonetic, orthographic, and cross-lingual encoding strategies. We evaluated language models on coded language detection, classification, and review rating prediction. Our results showed that even strong models frequently fail to detect coded language, and phonetic distortions contribute to downstream prediction errors. These findings highlight coded language as a persistent challenge for real-world NLP systems and motivate further work.

\section{Limitations\label{sec:limitation}}
Although our work represents an important first step toward understanding and modeling coded language, it has several limitations.

\begin{itemize}

\item 
First and most significantly, our data curation relied on an iterative bootstrapping process that was inherently constrained by what LLMs and existing lexicons could surface. 
If a coded expression never appeared in the seed dictionaries we used and was not recognized by the LLMs during bootstrapping, it was missed. 
This bias was difficult to avoid when constructing datasets for low-frequency linguistic phenomena and fundamentally limited coverage of rare, creative, or highly novel forms of coded language.

\item
Second, the dataset was temporally and geographically bounded.
Our collection endpoint was 2022, and the data focused on online reviews in the United States.
Because coded language evolves over time and varies across regions, platforms, and communities, newer forms or region-specific patterns emerging after this cutoff were not captured. 
In the long run, we plan to update \dataset to cover additional languages, genres, and time frames.

\item
Third, annotation and interpretation were inevitably shaped by the annotators' linguistic backgrounds and community knowledge. 
As a result, some coded expressions may have been missed or misinterpreted, particularly those that were highly personalized, culturally specific, or dependent on insider context.

\item
Fourth, the operational scope of coded language in this paper was informed by collective authorial decisions made during taxonomy development and data curation. 
We were confident that such cases were rare, given that the authors have professional training in linguistics and are native speakers of Chinese. 
However, it remained possible that a different group of researchers, drawing on alternative theoretical perspectives or community norms, might have made different judgments about which cases to include or exclude.

\item
Finally, our analysis prioritized identifying broad patterns and evaluating downstream impacts of coded language rather than constructing an exhaustive inventory. 
Accordingly, we did not claim to capture all possible coded expressions in online communication.

\end{itemize}


\section{Ethics Statement\label{sec:ethics}}
Our work analyzed naturally occurring coded language in online reviews to better understand how meaning is expressed beyond literal text. 
Some coded expressions are intentionally used to obscure meaning or to restrict interpretability to specific audiences. 
We recognized that, in some cases, decoding such expressions could run counter to the original communicative intent of individual writers. 
We are aware of and familiar with prior scholarship on ethical considerations in research using public online data, including discussions of participant perspectives and ethical tensions when analyzing social media content in public research contexts, such as studies of Twitter user perceptions of research ethics~\cite{doi:10.1177/2056305118763366} and systematic reviews of ethical questions in research using Reddit data~\cite{doi:10.1177/2056305118763366}.

To mitigate ethical concerns, we focused on a relatively low-risk domain, namely publicly available online reviews. 
All data were drawn from public sources and analyzed in aggregate, without attempts to infer, identify, or attribute sensitive information to specific individuals.

Our goal was not to expose or target individual users, but to study coded language as a linguistic and social phenomenon that plays an important role in human communication. 
By examining coded language, we aimed to support more faithful and equitable interpretation of authentic user communication in NLP systems, rather than to undermine or suppress user expression.




\section*{Acknowledgments}
We acknowledge support from the Linguistic Diversity Across the Lifespan Graduate Research Traineeship Program (NSF Grant No. 2125865) awarded to Ruyuan Wan, as well as the pilot-study funding from the Center for Language Science for this project. We also thank the anonymous reviewers, Dr. Kenton Murray, Dr. Mina Lee and her lab, the Cornell NLP Lab, Dr. Qunfang Wu, Lan Gao, and Lu Xian for their constructive feedback.

\bibliography{custom}

\appendix

\section{Appendix\label{sec:appendix}}
\subsection{Statistics Analysis}
\label{sec:stat}

We examine whether coded language is associated with user ratings in Google Reviews. Using a chi-square test of independence, we find a significant association between coded language and rating distribution ($p < 0.001$), with a moderate effect size (Cram\'er’s $V = 0.29$). A Mann--Whitney $U$ test further shows that reviews containing coded language have significantly lower rating ranks than non-coded reviews ($p < 0.001$). Consistent with these findings, an ordinal logistic regression indicates a negative association between coded language presence and rating ($\beta = -0.89$, $p < 0.001$), suggesting that coded language is more likely to appear in lower-rated reviews.


\begin{figure}
    \centering
    \includegraphics[width=1\linewidth]{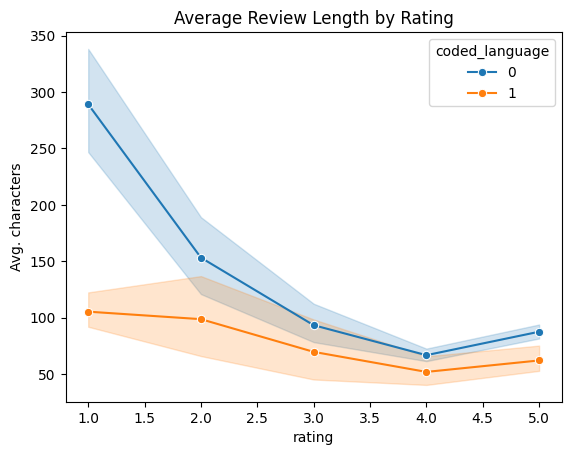}
    \caption{The average review length by rating, comparing coded and non-coded reviews.}
    \label{fig:len_rating}
\end{figure}

We further analyze the relationships among review length, coded language use, and rating scores. 
Because the review length distribution is highly skewed and heavy-tailed, we employ nonparametric statistical tests throughout.
We first examine the relationship between review length and rating across the full corpus. A Kruskal–Wallis test shows that the distributions of review lengths differ significantly across the five rating levels ($p < 0.001$). Consistent with this result, Spearman’s rank correlation reveals a weak but significant negative monotonic relationship between rating and review length ($\rho = -0.12, p < 0.001$), indicating that lower-rated (more negative) reviews tend to be longer overall.
We next test whether coded language use is associated with review length when aggregating across all ratings. A Mann–Whitney U test comparing coded and non-coded reviews shows no significant difference in length distributions ($p = 0.18$), suggesting that coded language is not globally associated with review length. However, when stratifying by rating, length differences emerge at the extremes: coded reviews are significantly shorter than non-coded reviews at both 1 and 5 ratings ($p < 0.001$), while differences at intermediate ratings (2–4) are not statistically significant. This pattern indicates that the length effect of coded language is conditional on sentiment intensity rather than a uniform stylistic property.
Therefore, negative reviews tend to be longer in general, while coded language is associated with more concise expression under extreme sentiment conditions, as also shown in Figure \ref{fig:len_rating}.

\subsection{Additional Coded Language Classification Results}
\label{sec:additional_classification}
Table~\ref{tab:combined_classification} reports per-class precision, recall, and F1 across GPT-5-mini, Gemini-2.5-Flash, Qwen2.5-7B-Instruct and 
Llama-3.1-8B-Instruct.

\begin{table*}[t]
\centering
\small
\setlength{\tabcolsep}{3.6pt}
\begin{tabular}{lcccccccccccc}
\toprule
& \multicolumn{3}{c|}{\textbf{GPT-5-mini}} 
& \multicolumn{3}{c|}{\textbf{Gemini-2.5-Flash}} 
& \multicolumn{3}{c|}{\textbf{Qwen2.5-7B-Instruct}} 
& \multicolumn{3}{c}{\textbf{Llama-3.1-8B-Instruct}} \\

\textbf{Coded Language Class} 
& \textbf{P}  & \textbf{R}  & \textbf{F$_1$}  
& \textbf{P}  & \textbf{R}  & \textbf{F$_1$}
& \textbf{P}  & \textbf{R}  & \textbf{F$_1$}  
& \textbf{P}  & \textbf{R}  & \textbf{F$_1$}  \\
\midrule

Ambiguous Homophone 
& 0.66 & 0.61 & 0.63 
& 0.57 & 0.61 & 0.59 
& 0.45 & 0.23 & 0.30 
& 0.25 & 0.27 & 0.26\\

Non-Lexical Homophone 
& 0.65 & 0.54 & 0.59 
& 0.68 & 0.73 & 0.70 
& 0.38 & 0.30 & 0.34 
& NaN  & 0.00 & NaN\\

Phonetic Substitution 
& 0.50 & 0.95 & 0.65 
& 0.51 & 0.97 & 0.67 
& 0.48 & 0.16 & 0.24 
& 0.38 & 0.45 & 0.41\\

Emoji Substitution 
& 0.75 & 0.25 & 0.38 
& 0.77 & 0.83 & 0.80 
& 0.10 & 0.08 & 0.09 
& 0.09 & 0.58 & 0.16\\

Cipher 
& 0.86 & 0.81 & 0.84 
& 0.78 & 0.85 & 0.81 
& 0.45 & 0.12 & 0.19 
& 0.04 & 0.13 & 0.06\\

Orthographic Substitution 
& 0.30 & 0.91 & 0.45 
& 0.46 & 0.97 & 0.63 
& 0.17 & 0.65 & 0.27 
& 0.02 & 0.85 & 0.05\\

Cross-Lingual Phonetic Encoding
& 0.05 & 1.00 & 0.10 
& 0.04 & 1.00 & 0.08 
& 0.09 & 0.57 & 0.16 
& 0.06 & 0.71 & 0.11\\

\bottomrule
\end{tabular}
\caption{Per-class coded language classification performance across models.}
\label{tab:combined_classification}
\end{table*}

\subsection{IPA-based Phonetic Analysis}
\label{app:IPA}
\begin{figure}[t]
\centering
\includegraphics[width=1\linewidth]{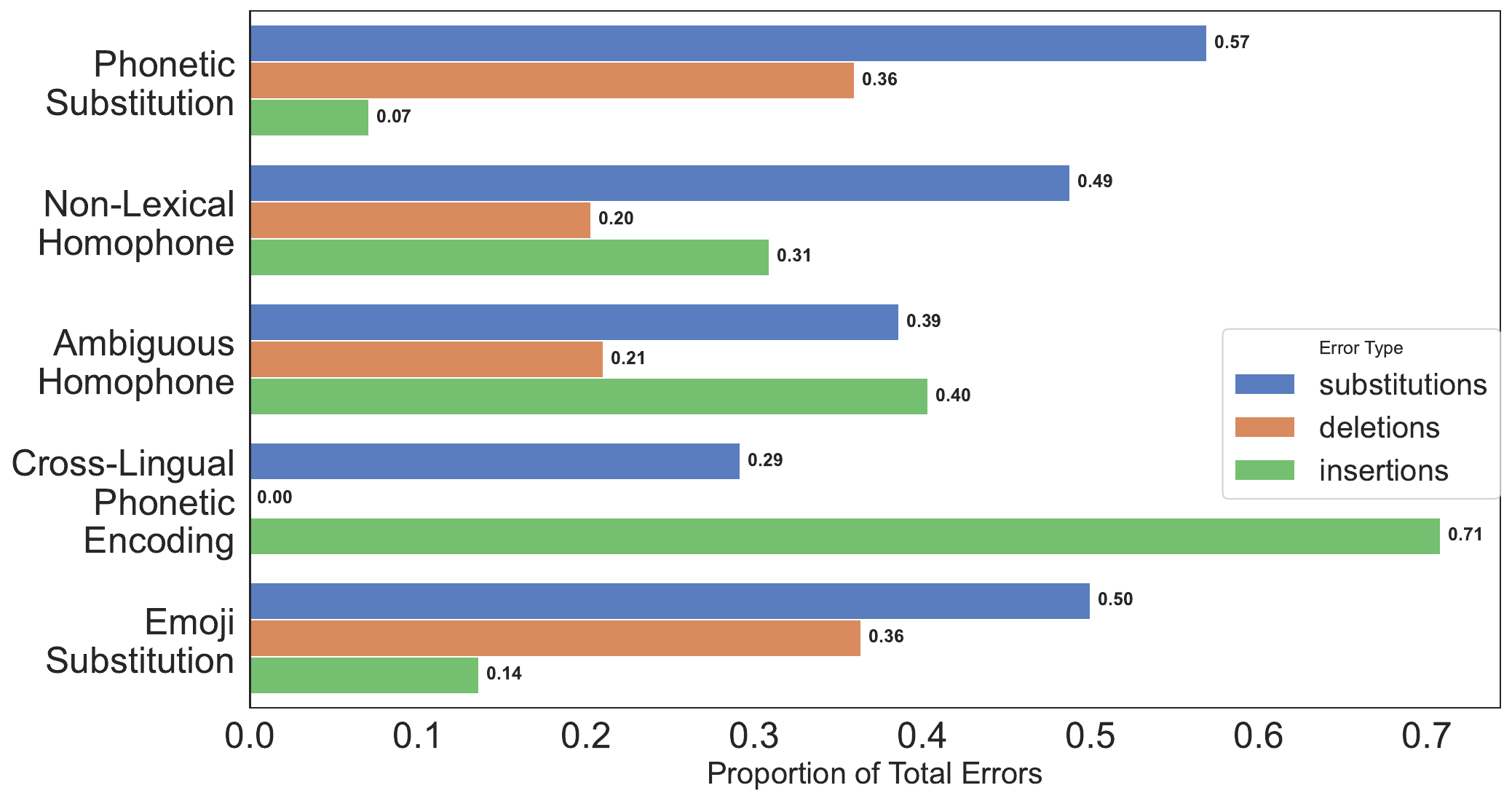}
\caption{The average error rate for substitutions, deletions, and insertions per class using IPA representations, ranked by the total number of errors per class.}
\label{fig:IPA}
\end{figure}

Figure \ref{fig:IPA} presents character error rate analyses computed using IPA representations, corresponding to the IPA-based results reported in the main paper.

\subsection{Prompt Templates}

\subsubsection{Coded Language Detection and Classification}

\begin{CJK}{UTF8}{gbsn}
\begin{verbatim}
You are annotating whether a review
contains coded language.
Definition:
"Coded language" refers to intentional
encoding or obfuscation of meaning.
The surface form and the intended hidden
meaning are different, and decoding
or inference is required to recover
the meaning.
IMPORTANT:
Not all socially opaque language is
coded language. Dialect, jargon, or 
multilingual text alone does NOT count 
as coded language unless encoding is
involved.
Coded language classes (multi-label
allowed):

1. ambiguous_homophone
  An existing word or phrase whose
  literal meaning differs from its
  hidden meaning.
  Examples:
  - 炒鸡 → 超级 (very)
  - 辣鸡 → 垃圾 (trash)
  - 稀饭 → 喜欢 (like)
  
2. non_lexical_homophone
  Homophonic characters or tokens 
  that do NOT form a valid lexical
  expression on the surface.
  Examples:
  - 灰常 (非常)
  - 菇凉 (姑娘)
  - 开森 (开心)
  
3. Phonetic Substitution
  Use of pinyin, phonetic symbols,
  or roman letters to represent 
  words or meanings instead of 
  full characters.
  Examples:
  - nb （牛逼）
  - tm （他妈）
  - 好ㄘ（好吃）
  - e度（呢度）
  
4. Emoji 
  Emoji replaces textual content and 
  conveys a hidden meaning DIFFERENT 
  from its literal emoji meaning.
  Coded examples:
  - [鸡emoji]贼 → 鸡贼
  - ①般 → 一般
  - 铁[汉堡emoji][汉堡emoji] → 铁憨憨
  Non-coded (DO NOT label as coded):
  - 值[得emoji] → (值得)
  - [满emoji]意 → 满意
  - 酱油[鸡emoji] → (酱油鸡)
  
5. orthographic
  Encoding through visual or orthographic
  similarity.
  Examples:
  - 爪ㄚ → MY
  - 艹 → 草
  - N0 G00D → NO GOOD
  
6. transliteration
  Phonetic transliteration across
  languages.
  Examples:
  - 古德 → good
  - 歪瑞古德 / 歪立古德 → very good
  - 纱头 → shuttle
  - ball ball → 求求
  
7. cipher
  The sentence has no clear literal
  semantic meaning and appears
  intentionally nonsensical or 
  cipher-like.
  Example:
  - qqQQaQ！你燃yfyi y f r
  
Exclusions (NOT coded language):
- Pure dialectal expressions without
encoding
- Pure multilingualism without
phonetic or orthographic substitution
- Emoji used literally without hidden
meaning
Your task:
- Decide whether coded language is 
present.
- If present, identify ALL applicable
coded language classes.
- Extract exact spans from the review 
that constitute coded language.
Return ONLY valid JSON with the 
following keys:
- label: 1 if coded language is 
present, else 0
- classes: list of coded language 
classes (empty if label=0)
- evidence: list of exact spans 
copied from the review 
(empty if label=0)
- short_reason: <= 20 words 
explaining the decision

Review:
<<<{text}>>>

\end{verbatim}
\end{CJK}

\subsubsection{Rating Prediction}
\paragraph{Original Reviews}
\begin{verbatim}
You will be given a review.
Estimate the user's intended star rating
on a 1–5 scale.

Rules:
- Output JSON ONLY.
- The JSON must have exactly one field:
"rating".
- The value of "rating" must be a number
between 1 and 5.
- Do not include any explanation or 
extra fields.

Review:
{text}
\end{verbatim}
\paragraph{Masked Reviews}
\begin{verbatim}
You will be given a review.
Estimate the user's intended star rating
on a 1–5 scale.

Note:
- The token [CODE] indicates masked 
coded language in the original review.
- Treat [CODE] as an unknown placeholder

Rules:
- Output JSON ONLY.
- The JSON must have exactly one field:
"rating".
- The value of "rating" must be a number 
between 1 and 5.
- Do not include any explanation or extra
fields.

Review:
{text}
\end{verbatim}
\paragraph{Decoded Reviews}
\begin{verbatim}
You will be given a review.
Estimate the user's intended star rating 
on a 1–5 scale.

Note:
- Some reviews may contain unusual
or nonsensical expressions.

Rules:
- Output JSON ONLY.
- The JSON must have exactly one field:
"rating".
- The value of "rating" must be a 
number between 1 and 5.
- Do not include any explanation or 
extra fields.

Review: 
{text}
\end{verbatim}

\end{document}